\documentclass[sigconf]{acmart}

\AtBeginDocument{%
  \providecommand\BibTeX{{%
    \normalfont B\kern-0.5em{\scshape i\kern-0.25em b}\kern-0.8em\TeX}}}

\setcopyright{rightsretained}
\copyrightyear{2021}
\acmYear{2021}
\acmDOI{}

\acmConference[SiKDD '22]{Ljubljana '22: Slovenian KDD Conference on Data Mining and Data Warehouses}{October, 2022}{Ljubljana, Slovenia}
\acmBooktitle{Ljubljana '22: Slovenian KDD Conference on Data Mining and Data Warehouses, October, 2022, Ljubljana, Slovenia}
\acmPrice{}
\acmISBN{}

\usepackage{listings}
\usepackage{xcolor}
\usepackage{adjustbox}
\usepackage{float}
\usepackage{color,soul}
\usepackage{aliascnt}
\usepackage{amsmath}
\usepackage{textgreek}
\usepackage{caption}
\usepackage{tikz-cd}
\usepackage{graphicx}
\usepackage{multirow}
\usepackage{hyperref}
\usepackage{academicons}
\restylefloat{table}

\newaliascnt{eqfloat}{equation}
\newfloat{eqfloat}{h}{eqflts}
\floatname{eqfloat}{Equation}
\newcommand*{\ORGeqfloat}{}
\let\ORGeqfloat\eqfloat
\def\eqfloat{%
  \let\ORIGINALcaption\caption
  \def\caption{%
    \addtocounter{equation}{-1}%
    \ORIGINALcaption
  }%
  \ORGeqfloat
}

\begin{document}

\title{Machine Beats Machine: Machine Learning Models to Defend Against Adversarial Attacks.}


\author{Jo\v{z}e M. Ro\v{z}anec}
\authornotemark[1]
\orcid{0000-0002-3665-639X}
\affiliation{%
  \institution{Jo\v{z}ef Stefan International Postgraduate School}
  \streetaddress{Jamova 39}
  \city{Ljubljana}
  \country{Slovenia}}
\email{joze.rozanec@ijs.si}

\author{Dimitrios Papamartzivanos}
\orcid{0000-0002-9471-5415}
\affiliation{%
  \institution{Ubitech Ltd}
  \streetaddress{Thessalias 8 and Etolias 10}
  \city{Chalandri, Athens}
  \country{Greece}}
\email{dpapamartz@ubitech.eu}

\author{Entso Veliou}
\orcid{0000-0001-9730-1720}
\affiliation{%
  \institution{Department of Informatics and Computer Engineering, University of West Attica}
  \streetaddress{Agiou Spyridonos Street}
  \city{Athens}
  \country{Greece}}
\email{eveliou@uniwa.gr}

\author{Theodora Anastasiou}
\orcid{0000-0002-9766-9338}
\affiliation{%
  \institution{Ubitech Ltd}
  \streetaddress{Thessalias 8 and Etolias 10}
  \city{Chalandri, Athens}
  \country{Greece}}
\email{tanastasiou@ubitech.eu}

\author{Jelle Keizer}
\orcid{0000-0002-8020-4120}
\affiliation{%
  \institution{Philips Consumer Lifestyle BV}
  \streetaddress{Oliemolenstraat 5}
  \city{Drachten}
  \country{The Neatherlands}}
\email{jelle.keizer@philips.com}

\author{Bla\v{z} Fortuna}
\orcid{0000-0002-8585-9388}
\affiliation{%
  \institution{Qlector d.o.o.}
  \streetaddress{Rov\v{s}nikova 7}
  \city{Ljubljana}
  \country{Slovenia}}
\email{blaz.fortuna@qlector.com}

\author{Dunja Mladeni\'{c}}
\orcid{0000-0003-4480-082X}
\affiliation{%
  \institution{Jo\v{z}ef Stefan Institute}
  \streetaddress{Jamova 39}
  \city{Ljubljana}
  \country{Slovenia}}
\email{dunja.mladenic@ijs.si}

\renewcommand{\shortauthors}{Ro\v{z}anec et al.}

\begin{abstract}
We propose using a two-layered deployment of machine learning models to prevent adversarial attacks. The first layer determines whether the data was tampered, while the second layer solves a domain-specific problem. We explore three sets of features and three dataset variations to train machine learning models. Our results show clustering algorithms achieved promising results. In particular, we consider the best results were obtained by applying the DBSCAN algorithm to the structured structural similarity index measure computed between the images and a white reference image.
\end{abstract}




\maketitle

\section{Introduction}


Artificial Intelligence (AI) solutions have penetrated the Industry 4.0 domain by revolutionizing the rigid production lines enabling innovative functionalities like mass customization, predictive maintenance, zero defect manufacturing, and digital twins. However, AI-fuelled manufacturing floors involve many interactions between the AI systems and other legacy Information and Communications Technology (ICT) systems, generating a new territory for malevolent actors to conquer. Hence, the threat landscape of Industry 4.0 is expanded unpredictably if we also consider the emergence of adversary tactics and techniques against AI systems and the constantly increasing number of reports of Machine Learning (ML) systems abuses based on real-world observations. In this context, Adversarial Machine Learning (AML) has become a significant concern in adopting AI technologies for critical applications, and it has already been identified as a barrier in multiple application domains. AML is a class of data manipulation techniques that cause changes in the behavior of AI algorithms while usually going unnoticed by humans. Suspicious objects misclassification in airport control systems~\cite{gota2020threat}, abuse of autonomous vehicles navigation systems ~\cite{kloukiniotis_countering_2022}, tricking of healthcare image analysis systems for classifying a benign tumor as malignant~\cite{ma2021understanding}, abnormal robotic navigation control~\cite{zhang2015towards} are only a few examples of AI models’ compromise that advocate the need for the investigation and development of robust defense solutions. 

Recently, the evident challenges posed by AML have attracted the attention of the research community, the industry 4.0, and the manufacturing domains~\cite{veliou2021}, as possible security issues on AI systems can pose a threat to systems reliability, productivity, and safety~\cite{becue2021artificial}. In this reality, defenders should not be just passive spectators, as there is a pressing need for robustifying AI systems to hold against the perils of adversarial attacks. New methods are needed to safeguard AI systems and sanitize the ML data pipelines from the potential injection of adversarial data samples due to poisoning and evasion attacks.

\begin{figure}[h]
\includegraphics[width=0.5\textwidth,keepaspectratio]{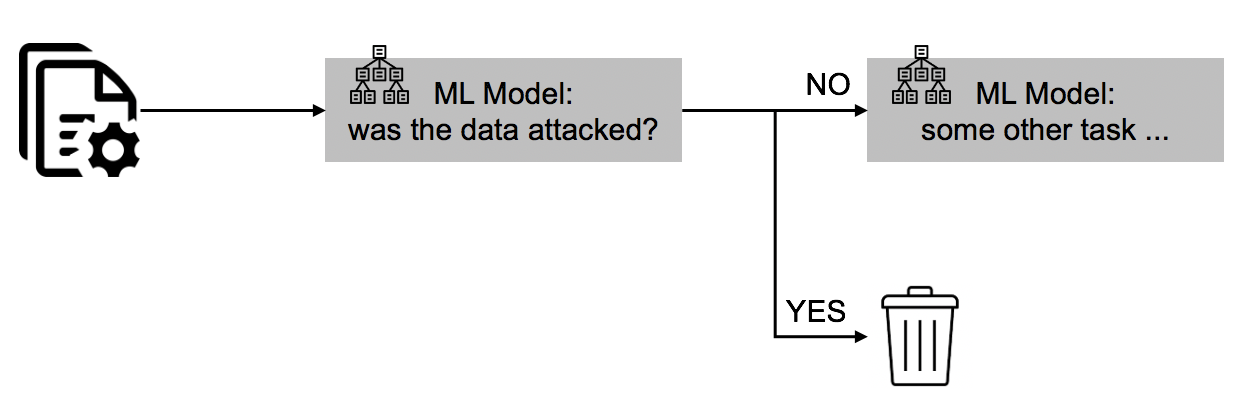}
\caption{Two-layered deployment of machine learning models can be used to prevent adversarial attacks.}
\label{CYBER:F:PIPELINE}
\end{figure}

We developed a machine learning model to address the abovementioned challenges, detecting whether the incoming images are adversarially altered. This enables a two-layered deployment of machine learning models that can be used to prevent adversarial attacks (see Fig. \ref{CYBER:F:PIPELINE}): (a) the first layer with models determining whether the data was tampered, and (b) a second layer that operates with regular machine learning models developed to solve particular domain-specific problems. We demonstrate our approach in a real-world use case from \textit{Philips Consumer Lifestyle BV}. This paper explores a diverse set of features and machine learning models to detect whether the images have been tampered for malicious purposes.

This paper is organized as follows. Section \ref{S:RELATED-WORK} outlines the current state of the art and related works, Section \ref{S:USE-CASE} describes the use case, and Section \ref{S:METHODOLOGY} provides a detailed description of the methodology and experiments. Finally, Section \ref{S:RESULTS} outlines the results obtained, while Section \ref{S:CONCLUSION} concludes and describes future work.

\section{Related Work}\label{S:RELATED-WORK}

AML attacks are considered a severe threat to AI systems, and, that is, the research community seeks new robust defensive methods. Image classifiers, those analyzed in this work, are the focal point of the vast majority of the AML literature, as those have been proved prone to noise perturbations. According to the literature, prominent solutions focus on denoising the image classifiers, training the target model with adversarial examples, known as adversarial training, or applying standalone defense algorithms. 

Yan et. al.~\cite{yan_towards_2022} proposed a new adversarial attack called Observation-based Zero-mean Attack, and they evaluated the robustness of various deep image denoisers. They followed an adversarial training strategy and effectively removed various synthetic and adversarial noises from data. In~\cite{pawlicki_preprocessing_2021}, pre-processing data defenses for image denoising are evaluated, highlighting the advantages of such approaches that do not require the retraining of the classifiers, which is a computationally intense task in computer vision. 

However, the robustness of adversarial training via data augmentation and distillation is advocated by the majority of the works in the domain. Specifically, Bortsova et al. ~\cite{bortsova_adversarial_2021} have focused on adversarial black-box settings, assuming that the attacker does not have full access to the target model as a more realistic scenario. They tuned their testbed to ensure minimal visual perceptibility of the attacks. The applied adversarial training dramatically decreased the performance of the designed attack. Hashemi and Mozaffari~\cite{Hashemi2021CNNAA} trained CNNs with perturbed samples manipulated by various transformations and contaminated by different noises to foster robustness using adversarial training.

On top of the above, several standalone solutions have been proposed. CARAMEL system in ~\cite{kyrkou_towards_2020} offered a set of detection techniques to combat security risks in automotive systems with embedded camera sensors. Hybrid approaches and more general alternatives intrinsically improve the robustness of AI models. A defensive Distillation mechanism against evasion attacks is proposed in \cite{DBLP:journals/corr/PapernotMWJS15} being able to reduce the effectiveness of adversarial sample creation from 95\% to less than 0.5\% on a studied DNN. Subset Scanning was presented in \cite{speakman2018subset} to give the ability to DNNs to recognize out-of-distribution samples.

\section{Use Case}\label{S:USE-CASE}
The Philips factory in Drachten, the Netherlands, is an advanced factory for mass manufacturing consumer goods (e.g., shavers, OneBlade, baby bottles, and soothers). Current production lines are often tailored for the mass production of one product or product series in the most efficient way. However, the manufacturing landscape is changing due to global shortages, manufacturing assets and components are becoming scarcer \cite{EUEF}, and a shift in market demand requires the production of smaller batches more often. To adhere to these changes, production flexibility, re-use of assets, and a reduction of reconfiguration times are becoming more critical for the cost-efficient production of consumer goods. One of the topics currently investigated within Philips
is quickly setting up automated quality inspections to make reconfiguring quality control systems faster and easier. Next to working on the technical challenges of doing this, safety and cyber-security topics are explored, aiming to implement AI-enabled automated quality systems with state-of-the-art defenses, the latter of which is the focus point discussed in this paper.

The dataset used contains images of the decorative part of a Philips shaver. This product is mass-produced and important for the visual appearance of the shavers. Next to that, the part is very close to or in direct contact with the user’s skin, where any deviations in its quality could impact shaver performance or even shaver safety. The dataset contains 1.194 images classified into two classes: (a) attacked with the Projected Gradient Descent attack \cite{deng2020universal}, and (b) not attacked. 

\section{Methodology}\label{S:METHODOLOGY}

\begin{figure}[h]
\includegraphics[width=0.45\textwidth,keepaspectratio]{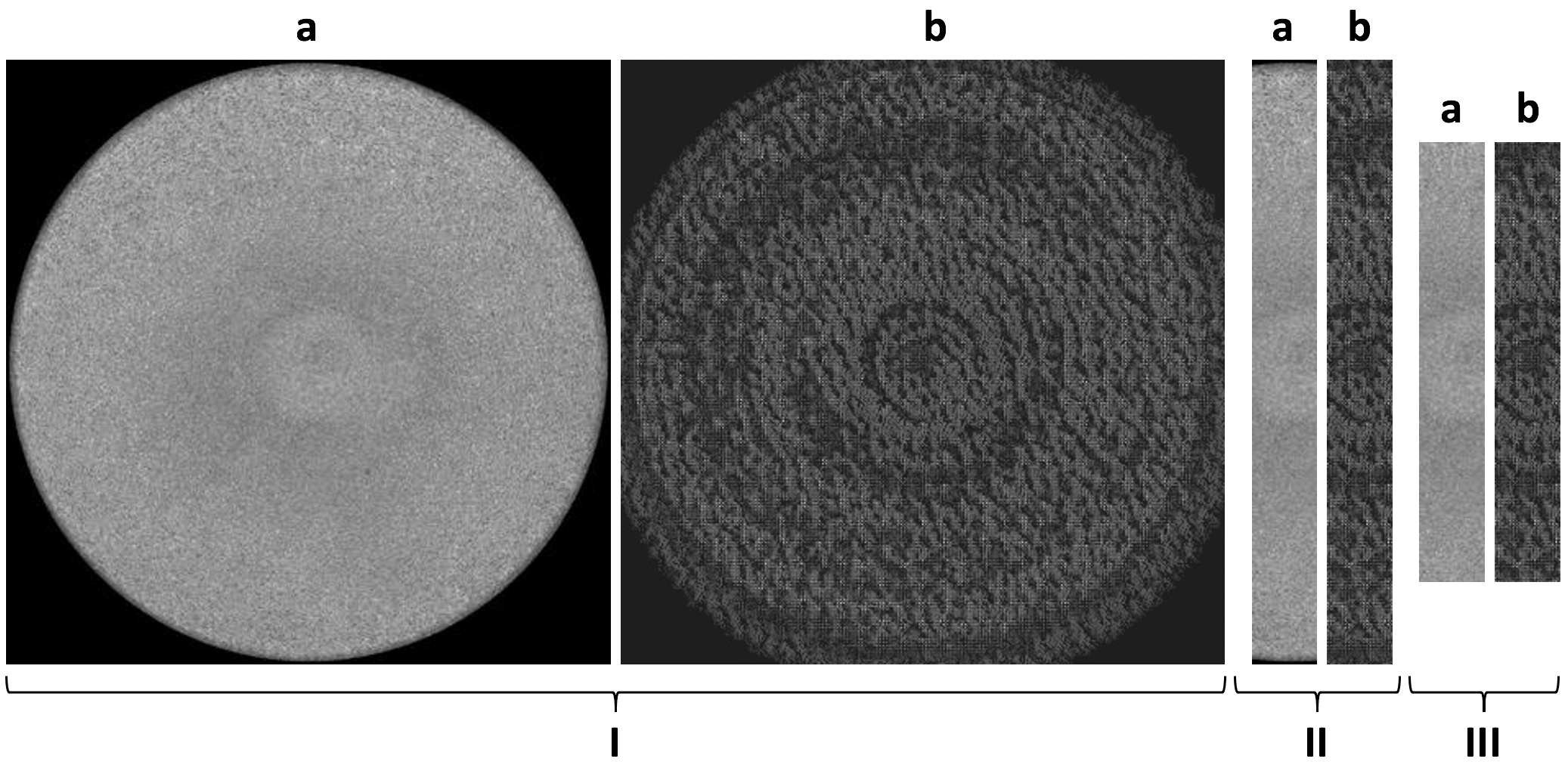}
\caption{Three sets of images: (a) indicates the original image, while (b) indicates the images attacked with the Projected Gradient Descent attack. The subsets I, II, and III indicate (I) the whole image, (II) cropped image (v1 (considering coordinates (160, 0, 200, 369))), and  cropped image (v2 - (considering coordinates (160, 50, 200, 319))).}
\label{CYBER:F:SAMPLE-IMAGES}
\end{figure}

We framed adversarial attack detection as a classification problem. We experimented with three kinds of features: (a) image embeddings (obtained from the  Average Pooling Layer of a pre-trained ResNet-18 model (\cite{he2016deep})), (b) histograms reflecting grayscale pixel frequencies (with pixel values extending between zero and 255), and (c) structural similarity index measure (SSIM) computed against a white image. While the embeddings provide information about the image as a whole, we considered the histograms and SSIM metric could be useful given the apparent difference between the original and perturbed images. Furthermore, we computed the features across three different datasets (see Fig. \ref{CYBER:F:SAMPLE-IMAGES} for sample images): (a) original set of images, (b) images cropped considering an image slice extending from top to bottom (coordinates (160, 0, 200, 369) - we name this dataset set "Cropped (v1)"), and (c) images cropped considering a slice of the central part of the image (coordinates (160, 50, 200, 319) - - we name this dataset set "Cropped (v2)"). By comparing the original image dataset against those obtained by slicing the central part, we sought to understand if the models' predictive power increased by looking at a specific area of the image rather than the whole.

We first trained three machine learning models: Catboost \cite{prokhorenkova2018catboost} with Focal Loss \cite{lin2017focal} (trained over 150 iterations, and considering a tree depth of ten, while evaluating the performance during training with the logloss metric), Logistic Regression (the dataset was scaled between zero and one, considering the train set, and transformed to ensure zero mean and unit variance), and KMeans (the dataset was transformed to ensure zero mean and unit variance, and the model initiated with random initialization and seeking to generate two clusters). We evaluated our experiments with a ten-fold stratified cross-validation (\cite{zeng2000distribution,kuhn2013applied}), using one fold for testing and the rest of the folds to train the model. Furthermore, to avoid overfitting, we performed a feature selection using the mutual information to evaluate the most relevant ones and select the \textit{top K} features, with $K=\sqrt{N}$, considering $N$ to be equal to the number of data instances in the train set \cite{hua2005optimal}. Finally, we measured our models' performance with a custom metric ($DP_{AUC ROC}$) that summarizes the discriminative power as computed from the area under the receiver operating characteristic curve (AUC ROC, see \cite{BRADLEY19971145}) (see Eq. \ref{E:DP-AUROC}). The metric ranges from zero (no discriminative power) to one (perfect discriminative power) and it preserves the AUC ROC desirable properties of being threshold independent and invariant to \textit{a priori} class probabilities.

\begin{eqfloat}
\begin{equation}\label{E:DP-AUROC}
    DP_{AUC_ROC} = 2 \cdot \lvert(0.5 - AUC ROC)\rvert 
\end{equation}
\end{eqfloat}

Based on the good results obtained in the clustering setting, we decided to conduct additional experiments, running the DBSCAN algorithm \cite{ester1996density} over all existing data. The advantage of such an algorithm is that it can estimate the clusters with no prior information regarding the number of expected clusters. Therefore, if working well, it would be useful to generalize the approach toward detecting new cyberattacks where no labeled data exists yet. We consider such a characteristic to be fundamental to production environments. For the models resulting from the three abovementioned datasets, we measured the estimated number of clusters, estimated number of noise points, homogeneity (whether the clusters contain only samples belonging to a single class), completeness (whether all the data points members of a given class are elements of the same cluster), V-measure (harmonic mean between homogeneity and completeness), adjusted Rand index (similarity between clusterings obtained by the proposed and random models), and the Silhouette Coefficient (estimates the separation distance between the resulting clusters). We ran the DBSCAN algorithm measuring the distance between clusters with the Euclidean distance, considering the maximum distance between two samples for one to be considered as in the neighborhood of the other to be 0,3. Furthermore, we considered that at least ten samples in a neighborhood were required for a point to be considered as a core point.

\section{Results and Analysis}\label{S:RESULTS}

\begin{table}[ht!]
\resizebox{0.45\textwidth}{!}{
\begin{tabular}{|ll|r|r|r|}
\hline
\multicolumn{2}{|c|}{\textbf{Model}} & \multicolumn{1}{l|}{\textbf{Catboost}} & \multicolumn{1}{l|}{\textbf{KMeans}} & \multicolumn{1}{l|}{\textbf{Logistic regression}} \\ \hline
\multicolumn{1}{|l|}{\multirow{3}{*}{\textbf{Embeddings}}} & \textbf{Original image} & 0.0167 & \textbf{1.0000} & \textit{0.0228} \\ \cline{2-5} 
\multicolumn{1}{|l|}{} & \textbf{Cropped (v1)} & \textit{0.0014} & \textbf{1.0000} & 0.0003 \\ \cline{2-5} 
\multicolumn{1}{|l|}{} & \textbf{Cropped (v2)} & 0.0181 & \textbf{1.0000} & \textit{0.0213} \\ \hline
\multicolumn{1}{|l|}{\multirow{3}{*}{\textbf{SSIM}}} & \textbf{Original image} & 0.0152 & \textbf{1.0000} & \textit{0.0184} \\ \cline{2-5} 
\multicolumn{1}{|l|}{} & \textbf{Cropped (v1)} & \textit{0.0008} & \textbf{1.0000} & 0.0004 \\ \cline{2-5} 
\multicolumn{1}{|l|}{} & \textbf{Cropped (v2)} & 0.0179 & \textbf{1.0000} & \textit{0.0195} \\ \hline
\multicolumn{1}{|l|}{\multirow{3}{*}{\textbf{Histograms}}} & \textbf{Original image} & 0.0016 & \textbf{1.0000} & \textit{0.0030} \\ \cline{2-5} 
\multicolumn{1}{|l|}{} & \textbf{Cropped (v1)} & 0.0003 & \textbf{1.0000} & \textit{0.0011} \\ \cline{2-5} 
\multicolumn{1}{|l|}{} & \textbf{Cropped (v2)} & 0.0018 & \textbf{1.0000} & \textit{0.0031} \\ \hline
\end{tabular}
\caption{Results obtained across classification experiments. We measure models' performance with Eq. \ref{E:DP-AUROC}. \textbf{Best results are bolded}, \textit{second-best are italicized}. \label{T:RESULTS-AUCROC}}}
\end{table}

We present the results obtained in our classification experiments in Table \ref{T:RESULTS-AUCROC}. We found the KMeans models achieved perfect discrimination in all cases, while the second-best model was the Logistic regression, which had second-best results in all but two cases. Nevertheless, the Logistic regression and the Catboost models achieved a low discriminative power, almost unable to distinguish between tampered and non-tampered images. Regarding the features, we found that the best average performance was obtained when training the models on the \textit{Cropped (v2)} dataset, followed by those trained on the whole images.

\begin{table*}[ht!]
\resizebox{\textwidth}{!}{
\begin{tabular}{|l|rrr|rrr|rrr|}
\hline
\multicolumn{1}{|c|}{\multirow{2}{*}{}} & \multicolumn{3}{c|}{\textbf{Embeddings}} & \multicolumn{3}{c|}{\textbf{SSIM}} & \multicolumn{3}{c|}{\textbf{Histograms}} \\ \cline{2-10} 
\multicolumn{1}{|c|}{} & \multicolumn{1}{l|}{\textbf{Original image}} & \multicolumn{1}{l|}{\textbf{Cropped (v1)}} & \multicolumn{1}{l|}{\textbf{Cropped (v2)}} & \multicolumn{1}{l|}{\textbf{Original image}} & \multicolumn{1}{l|}{\textbf{Cropped (v1)}} & \multicolumn{1}{l|}{\textbf{Cropped (v2)}} & \multicolumn{1}{l|}{\textbf{Original image}} & \multicolumn{1}{l|}{\textbf{Cropped (v1)}} & \multicolumn{1}{l|}{\textbf{Cropped (v2)}} \\ \hline
Number of clusters & \multicolumn{1}{r|}{3} & \multicolumn{1}{r|}{1} & 1 & \multicolumn{1}{r|}{\textbf{2}} & \multicolumn{1}{r|}{\textbf{2}} & \textbf{2} & \multicolumn{1}{r|}{1} & \multicolumn{1}{r|}{1} & 1 \\ \hline
Number of noise points & \multicolumn{1}{r|}{1010} & \multicolumn{1}{r|}{794} & 887 & \multicolumn{1}{r|}{1} & \multicolumn{1}{r|}{\textbf{0}} & 1 & \multicolumn{1}{r|}{621} & \multicolumn{1}{r|}{603} & 606 \\ \hline
Homogeneity & \multicolumn{1}{r|}{0.1770} & \multicolumn{1}{r|}{0.4550} & 0.3170 & \multicolumn{1}{r|}{\textbf{1.0000}} & \multicolumn{1}{r|}{\textbf{1.0000}} & \textbf{1.0000} & \multicolumn{1}{r|}{0.8550} & \multicolumn{1}{r|}{0.9290} & 0.9150 \\ \hline
Completeness & \multicolumn{1}{r|}{0.2090} & \multicolumn{1}{r|}{0.4940} & 0.3860 & \multicolumn{1}{r|}{\textit{0.9910}} & \multicolumn{1}{r|}{\textbf{1.0000}} & \textit{0.9910} & \multicolumn{1}{r|}{0.8560} & \multicolumn{1}{r|}{0.9290} & 0.9150 \\ \hline
V-measure & \multicolumn{1}{r|}{0.1920} & \multicolumn{1}{r|}{0.4740} & 0.3480 & \multicolumn{1}{r|}{\textit{0.9960}} & \multicolumn{1}{r|}{\textbf{1.0000}} & \textit{0.9960} & \multicolumn{1}{r|}{0.8550} & \multicolumn{1}{r|}{0.9290} & 0.9150 \\ \hline
Adjusted Rand index & \multicolumn{1}{r|}{0.0710} & \multicolumn{1}{r|}{0.4350} & 0.2540 & \multicolumn{1}{r|}{\textit{0.9980}} & \multicolumn{1}{r|}{\textbf{1.0000}} & \textit{0.9980} & \multicolumn{1}{r|}{0.9020} & \multicolumn{1}{r|}{0.9600} & 0.9500 \\ \hline
Silhouette coefficient & \multicolumn{1}{r|}{0.0750} & \multicolumn{1}{r|}{0.4310} & 0.2660 & \multicolumn{1}{r|}{0.8980} & \multicolumn{1}{r|}{\textbf{0.9590}} & \textit{0.9070} & \multicolumn{1}{r|}{0.8330} & \multicolumn{1}{r|}{0.8970} & 0.8800 \\ \hline
\end{tabular}
\caption{Results obtained across clustering experiments. \textbf{Best ones are bolded}, \textit{second-best are italicized}. \label{T:RESULTS-CLUSTERING}}}
\end{table*}

When running the DBSCAN algorithm (see results in Table \ref{T:RESULTS-CLUSTERING}), we found the best results were obtained considering the SSIM measure. Furthermore, using the SSIM issued excellent results in all cases. The best ones were obtained considering the \textit{Cropped (v1) dataset}, while the second-best was achieved with the \textit{Cropped (v2) dataset}. Using the SSIM only, the DBSCAN algorithm was able to correctly group the instances into two groups and misclassified at most a single instance. However, the performance achieved either with embeddings or histograms was not satisfactory. When considering histogram features, the DBSCAN algorithm was not able to discriminate between instances, creating a single cluster. On the other hand, when considering embeddings, DBSCAN created three clusters that issued a bad performance, considering most of the points to be noisy. We, therefore, conclude that the only promising results were those obtained considering the SSIM. Nevertheless, we consider further research is required to understand whether this kind of feature can be useful across a wide range of attacks and in the real-world. SSIM provides metadata describing the images. Given high-quality attacks aim to reduce the visual footprint on the images, it remains an open question to which extent can the SSIM capture weak footprints and therefore enable similar discriminative capabilities on machine learning models.

\section{Conclusion}\label{S:CONCLUSION}
In this work, we explored multiple sets of features and machine learning models to determine whether an image has been tampered with for the purpose of an adversarial attack. While the difference between attacked and non-attacked images is evident to the human eye, it is not to the machine learning algorithms. We found that the Catboost and Logistic regression models could almost not discriminate between both cases. On the other hand, the clustering algorithms (KMeans and DBSCAN) had a stronger performance. While the KMeans models did so perfectly, regardless of the features, the DBSCAN model only performed well using the SSIM. We consider the strength of such a model the fact that no \textit{a priori} information regarding the classes is required, therefore saving the annotation effort and providing greater flexibility towards future adversarial attacks. Our future research will focus on testing a wider range of cyberattacks while ensuring the attack will not be discernable to the human eye.

\begin{acks}
This work was supported by the Slovenian Research Agency and the European Union’s Horizon 2020 program project STAR under grant agreement number H2020-956573.
\end{acks}

\bibliographystyle{ACM-Reference-Format}
\bibliography{main}

\end{document}